\documentclass{article}

\usepackage[preprint,nonatbib]{nips_2018}

\usepackage[utf8]{inputenc} 
\usepackage[T1]{fontenc}    
\usepackage{hyperref}       
\usepackage{url}            
\usepackage{booktabs}       
\usepackage{amsfonts}       
\usepackage{nicefrac}       
\usepackage{microtype}      

\usepackage{subfig}
\usepackage{graphicx}

\title{Skin lesion segmentation using a U-Net and\\
  good training strategies}

\author{
  Frederico Guth ~ and ~ Teofilo E. deCampos\thanks{http://cic.unb.br/\~{}teodecampos}\\
  Departamento de Ciência da Computação,\\
  Universidade de Brasília (UnB), Bras\'{\i}lia-DF, Brazil, CEP 70910-900 \\
  \texttt{fredguth@fredguth.com}, \texttt{t.decampos@oxfordalumni.org}
}

\begin{document}

\maketitle

\begin{abstract}
  In this paper we approach the problem of skin lesion segmentation using a
  convolutional neural network based on the U-Net architecture.
  We present a set of training strategies that had a significant impact
  on the performance of this model.
  We evaluated this method on the ISIC Challenge 2018 - Skin Lesion
  Analysis Towards Melanoma Detection, obtaining threshold Jaccard index of 77.5\%.
\end{abstract}

\section{Introduction}

According to the World Health Organization, between 2 and 3 million
non-melanoma skin cancers and 132,000 melanoma skin cancers occur
globally each year \cite{who}. Despite representing less than 6.5\% of
all skin cancers, melanomas are the most dangerous type, accounting
for approximately 75\% of all skin cancer related deaths
\cite{who,nature}.

Early detection is critical to increase survival expectancy and visual
inspection still is the most common diagnostic technique.

Deep convolutional neural networks (CNNs) already exceed human
performance in visual classification~\cite{fei}.
In some areas of oncology, such as histological image analysis, CNNs
have also proven to match the performance of experts,
e.g.\ \cite{veta_etal_mia2015}.  In an attempt to improve the
scalability of diagnostic expertise, CNNs have been developed to
locate and classify skin cancers in images with dermatologist-level
accuracy \cite{nature}.

Dermoscopy is a technique for examination of skin lesions that, with
proper training, increase diagnostic accuracy from 60\% (unaided expert
visual inspection) to 75\%-84\%~\cite{isic}. The International Skin
Imaging Collaboration (ISIC) has a large-scale publicly accessible
dataset of more than 20,000 dermoscopy images and host an annual
benchmark challenge on dermoscopic image analysis since 2016.  The
challenge comprises 3 tasks of lesion analysis: Segmentation,
Dermoscopic feature extraction and Classification.  In this paper, we
present results on Segmentation, identifying the lesion region in 
dermoscopic images. To our knowledge, we are the first to apply,
for this task, an architecture based on U-Net with a combination
of recent training strategies. 

\section{The model: U-Net34}
In this paper, we employed U-Net34, which combines insights from both
U-Net and Resnet.

Introduced in 2015, U-Net is an encoder-decoder architecture designed for biomedical image segmentation~\cite{ronneberger_etal_UNet_miccai2015},
with has later been employed for other image segmentation problems
as well, such as satellite image analysis~\cite{iglovikov}.
In a U-Net, the output is an image with the same dimension of the input,
but with one channel (in the case of binary segmentation problems).
The encoder path is a typical CNN, where each down-sampling step doubles
the number of feature channels.
What makes this architecture unique is the decoder path, where each
up-sampling step input is a concatenation of the output of the previous
step with the output of the corresponding (same height) down-sampling step.
This strategy enables precise localization with a very simple network. 

Resnet is a very successful architecture in several visual classification
tasks~\cite{he}.
It mitigates the degradation problem that happens when very deep
networks starts converging.
Instead of learning a direct mapping $H(x) = y$, it learns the residual
function  $F(x) = H(x)-x$, which can be re-framed into
$H(x) = F(x)+x = y$, where $F(x)$ is a stack of non-linear layers and
$x$ is the identity function(input=output).
The formulation of $F(x)+x$ can be implemented by feed-forward neural
networks with ``shortcut connections''.
Resnet34, specifically, is composed of an initial convolutional layer,
16 blocks of 2 layers and a final fully connected layer.

The U-Net34 architecture uses a pretrained Resnet34 model as a
U-Net encoder path \cite{fastai}. First, every step from the adaptive
pooling onwards is removed, keeping only Resnet backbone.
Then we save the output of results of the initial layer, $3^{rd}$, $8^{th}$,
and $14^{th}$ blocks (of 16 in total).
During the up-sampling we concatenate the output of those with
the outputs of up-sampling steps.
We used Adam optimizer and Binary Cross Entropy with Logits as the loss function.

\section{Training strategies}
\label{sec:seg_training}

\subsection{Inductive transfer via fine tuning}
Since U-Net34 uses Resnet34, and a pre-trained Resnet34 model is available
for the ImageNet classification task~\cite{russakovsky_etal_ILSVRC_ijcv2015}, we use it as starting point
for the optimization of the encoding layers.

\subsection{Pyramid transfer} 
Since U-Net is a fully convolutional network, it should
(in theory) not be limited by a fixed input/output resolution.
This enables the use of insights inspired on image pyramids,
such that the network is first trained with low resolution
data and the convolutional layers learn contextual information.
Next, the network is progressively fine tuned with higher resolution
data and the convolutional layers learn fine details.

In our work, we first train the model with $128\times128$ images
and transfer this learning to train the same model with images with
$256\times256$ images. We would suggest using the same strategy to go from
$256\times256$ to $512\times512$ images, though this can be costly
in terms of GPU memory usage, which turned out to be and issue
for our low budget machine.

\subsection{Learning rate schedule}
\begin{figure}
  \centering
  \subfloat[Learning Rate vs Validation Loss used for Learning Rate Optimization.]{\includegraphics[width=.485\columnwidth]{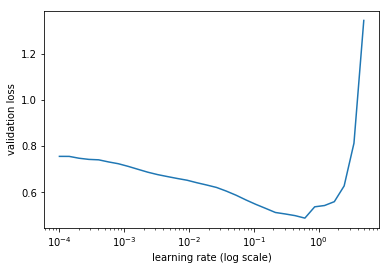}\label{lrfind}}
  \hfill
  \subfloat[Iterations vs Learning Rate of the STLR schedule.]{\includegraphics[width=.485\columnwidth]{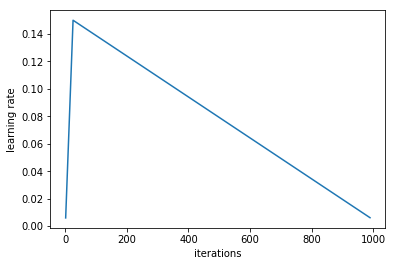}\label{sconv}}\hfil
\caption{Learning rate optimization and schedule.}\label{lr_find_chart}
\end{figure}

Our training process followed this procedure:
\begin{enumerate}
  \item Freeze the first layer group.
  \item \label{lr_find}Define the optimal learning rate with the method proposed by \cite{leslie} and implemented by \cite{fastai}, where one batch is trained with different learning rates, starting at very low and linearly increasing it at every iteration. Plotting a chart of the learning rate versus loss (see figure \ref{lrfind}) and choose the learning
    rate with the steepest downwards gradient on the validation loss.
  \item \label{superconvergence}Use the 1 cyclical learning rate policy (figure \ref{sconv}), also proposed by \cite{leslie}, to obtain training convergence in only 30 epochs (superconvergence). More specifically, we used the Slanted Triangular Learning Rate strategy (SLTR) \cite{howard_ruder_acl2018}.
  \item Unfreeze the model, keeping only the batch normalization layers frozen, and repeat steps \ref{lr_find} and \ref{superconvergence}.
\end{enumerate}

\subsection{Loss function}
It would be advisable to use a loss function more similar to the evaluation criteria. As the Jaccard index is not differentiable,
one could use a soft Jaccard variation\cite{iglovikov}.
However, in our preliminary trials the implemented soft Jaccard did not improve over the Binary Cross Entropy with Logits loss function and we decided to use the later. 

\subsection{Model selection and ensemble}
We evaluated two strategies for segmentation:
\begin{itemize}
\item \emph{BestDice}: This strategy just predicts the input with the model that presented the best dice index on the validation of our split out training set.
\item \emph{Ensemble}: We used the 3-folds of our training dataset and trained with BestDice model and ensemble to give a prediction.
\end{itemize}

\subsection{Data and Augmentation}
We used ``ISIC 2018: Skin Lesion Analysis Towards Melanoma Detection''
grand challenge datasets \cite{codella, ham} and no additional external
data. The Segmentation dataset comprises 2597 training images and 101
validation images acquired with a variety of dematoscope types, from
different anatomic sites, from sample of patients of different
institutions. There are more benign lesions than malignant, but an
over-representation of malignancies. Mask images are encoded as
gray-scale 8-bit PNGs, where each pixel is either 0, background, or 255,
lesion.

All images were first re-sized to $128\times128$ pixels, $256\times256$ and $512\times512$; and preprocessed to adjust color balance.
Random transformations on input images to augment the dataset were made: dihedral transformation, rotation (up to 44 degrees), zooming (up to 1.05), flipping and random lighting changes.
The official training dataset was then split out in 3-folds of training and validation datasets.

\section{Results and conclusion}

The best result on our validation set was obtained using the Ensemble strategy.  It achieves a 85.39\% Jaccard index and 78.43\% Threshold Jaccard index (with cut at 65\%).
The BestDice strategy achieved an online score of 75.5\% with the official validation set.
The top ranked participant in 2017 achieved an average Jaccard Index of 76.5\%, which should be compared with our 85.39\% score.

Visually the best segmentations are almost identical to the ground truth, but we can learn even more from our mistakes.
By analyzing the worse segmentations there are cases where, as non specialists, is hard to judge if the algorithm was wrong or the ground truth flawed. There are cases where our algorithm got confused by the pen marker or the glass used by the doctor; and it is clear that in general it does not do a good job when the lesion is small relative to the overall image. 

To conclude, we proposed to use a U-Net architecture for
segmentation of skin lesions. Our main contribution was the
combination of this architecture with a number of
training strategies, most of them are quite recent. These strategies
enabled us to use a relatively simple end-to-end network to generate
finely detailed segmentation results.

\medskip

\small
\bibliographystyle{plain}

\clearpage
\section{Supplementary material: qualitative results}
\begin{figure}[hbt]
\centering
\subfloat[One of many good results obtained by our method. \label{good}]{\includegraphics[width=.9\columnwidth]{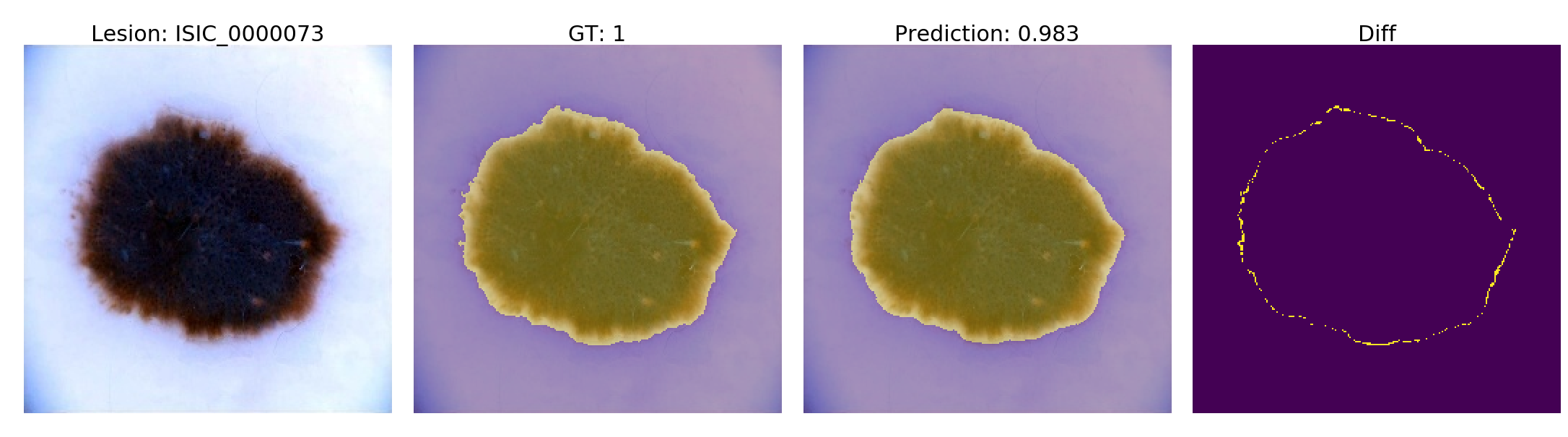}}\hfil
\subfloat[Failure cases.]{\includegraphics[width=.9\columnwidth]{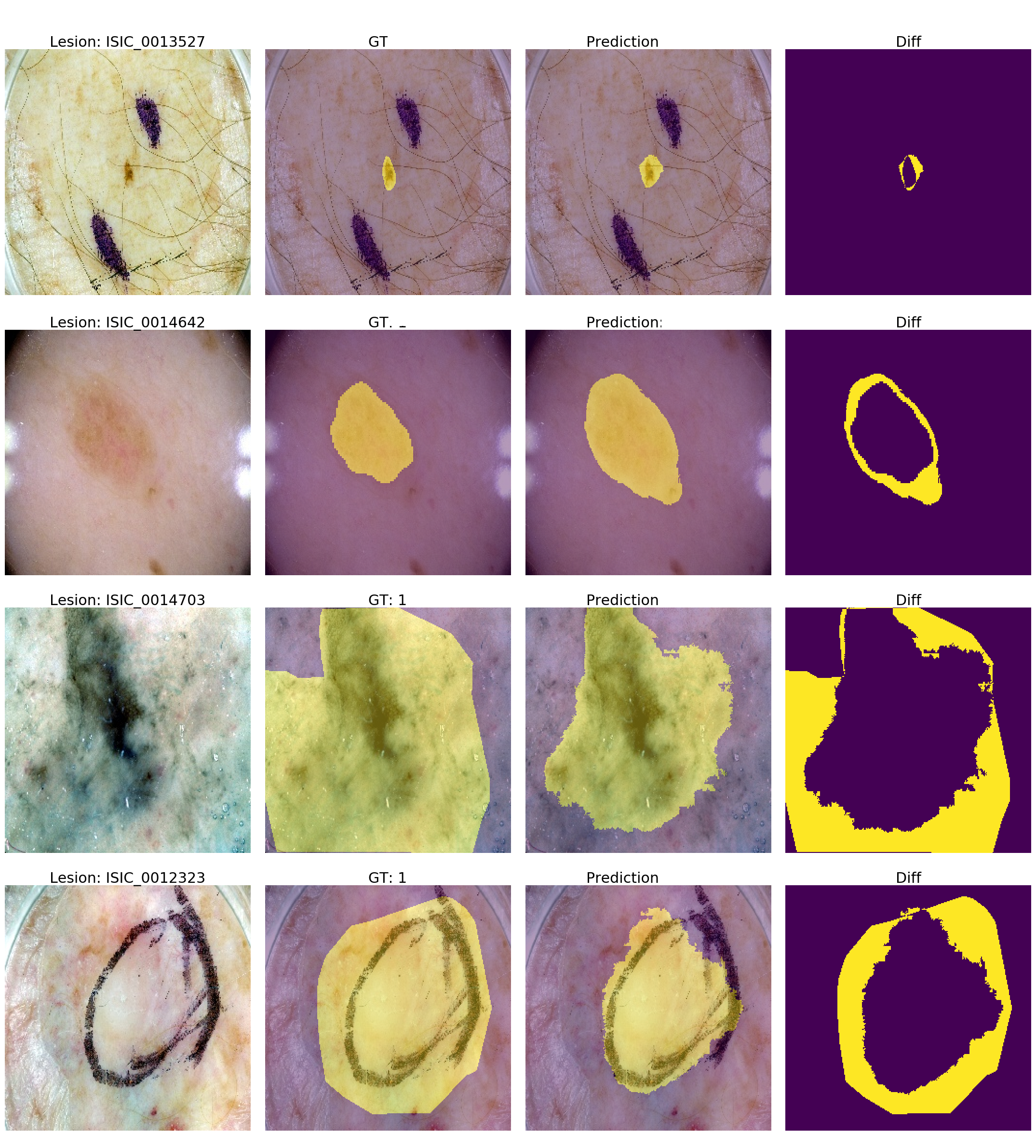}}
\caption{Qualitative assessment of segmentation results.}\label{result_samples}
\end{figure}

\end{document}